\documentclass[10pt,twocolumn,letterpaper]{article}

\usepackage[pagenumbers]{cvpr} 

\usepackage{graphicx}
\usepackage{amsmath}
\usepackage{amssymb}
\usepackage{booktabs}
\usepackage{multirow}
\usepackage{algorithm, algorithmic}

\DeclareFixedFont{\myfont}{U}{fplmbb}{m}{n}{10pt}
\usepackage[pagebackref,breaklinks,colorlinks]{hyperref}

\usepackage[capitalize]{cleveref}
\crefname{section}{Sec.}{Secs.}
\Crefname{section}{Section}{Sections}
\Crefname{table}{Table}{Tables}
\crefname{table}{Tab.}{Tabs.}

\def\cvprPaperID{3804} 
\def\confName{CVPR}
\def\confYear{2022}

\begin{document}




\title{Domain Generalization via Shuffled Style Assembly for Face Anti-Spoofing}





\author{
Zhuo Wang$^1$~~~~Zezheng Wang$^{2}$\thanks{\ denotes the corresponding author.}~~~~Zitong Yu$^{3}$~~~~Weihong Deng$^{1*}$\\
Jiahong Li$^{2}$~~~~Tingting Gao$^2$~~~~Zhongyuan Wang$^2$\\[1mm]
\normalsize{$^1$Beijing University of Posts and Telecommunications~~~~$^2$Kuaishou Technology~~~~$^3$CMVS, University of Oulu}\\
{\tt\small \{wz2019, whdeng\}@bupt.edu.cn~~~~zitong.yu@oulu.fi}\\
{\tt\small\{wangzezheng, lijiahong, wangzhongyuan\}@kuaishou.com~~~~tinagao2019@gmail.com}
}

\maketitle

\begin{abstract}
   With diverse presentation attacks emerging continually, generalizable face anti-spoofing (FAS) has drawn growing attention. Most existing methods implement domain generalization (DG) on the complete representations. However, different image statistics may have unique properties for the FAS tasks. In this work, we separate the complete representation into content and style ones. A novel \textbf{S}huffled \textbf{S}tyle \textbf{A}ssembly \textbf{N}etwork (SSAN) is proposed to extract and reassemble different content and style features for a stylized feature space. Then, to obtain a generalized representation, a contrastive learning strategy is developed to emphasize liveness-related style information while suppress the domain-specific one. Finally, the representations of the correct assemblies are used to distinguish between living and spoofing during the inferring. On the other hand, despite the decent performance, there still exists a gap between academia and industry, due to the difference in data quantity and distribution. Thus, a new large-scale benchmark for FAS is built up to further evaluate the performance of algorithms in reality. Both qualitative and quantitative results on existing and proposed benchmarks demonstrate the effectiveness of our methods. The codes will be available at \textcolor{magenta}{https://github.com/wangzhuo2019/SSAN}.
\end{abstract}


\section{Introduction}
\label{sec:intro}

\begin{figure}
	\begin{center}
		\includegraphics[width=7cm]{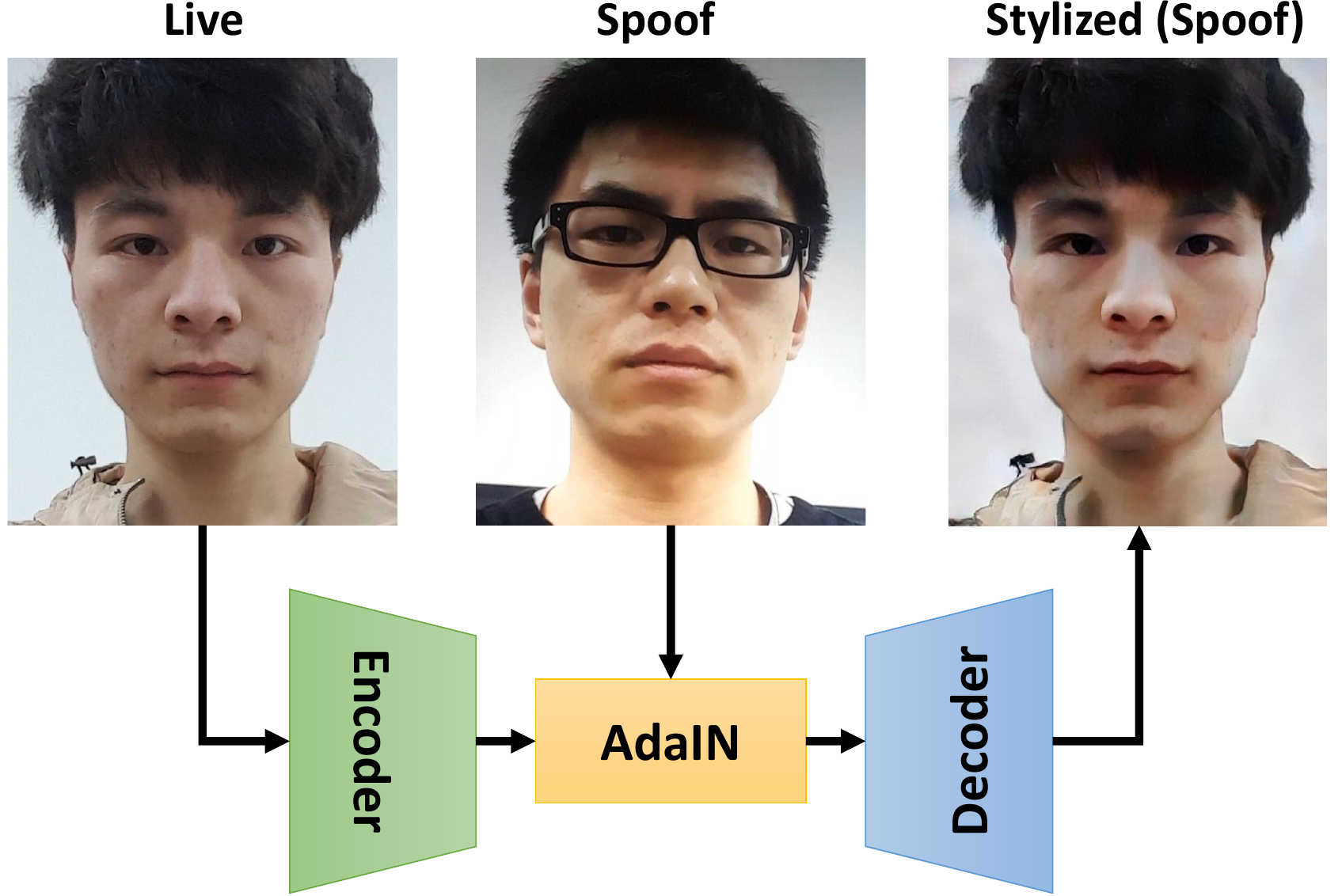}
	\end{center}
	\vspace{-17pt}
	\caption{The illustration of style transfer using the method of \cite{huang2017arbitrary} when live face as content input and spoof face as style input. }
	\label{fig:fig_head}
	\vspace{-8pt}
\end{figure}



As the most successful computer vision technology, face recognition (FR) \cite{wang2018deep,deng2019arcface}  has been widely employed in different application scenarios, such as mobile access control and electronic payments. Despite great success, FR systems may still suffer from presentation attacks (PAs), including print attacks, video replay, and 3D masks. To tackle these issues, a series of face anti-spoofing (FAS) methods have been proposed, from hand-craft descriptors based methods \cite{de2012lbp,patel2016secure} to deep representation based ones \cite{yu2021deep,yang2014learn,yu2020searching,yu2021dual,9730902}. 

The previous FAS methods have achieved promising performance in intra-domain scenarios, but may encounter dramatic degradation under the cross-domain settings. The major reason behind this lies in the conflict between the limitations of training data and the capability of networks \cite{he2016deep,liu2018learning,yu2020auto}, which makes the models trapped in dataset bias \cite{torralba2011unbiased} and leads to poor generalization toward new domains. To address this problem, domain adaptation (DA) techniques \cite{li2018unsupervised,wang2019improving} are used to alleviate the discrepancy between source and target domains by using unlabeled target data. However, in most real-world FAS scenarios, it is inefficient to collect sufficient unlabeled target data for training. 

Thus, domain generalization (DG) methods are proposed to generalize well on the unseen target domain, which can be coarsely classified into three categories: learning a common feature space \cite{shao2019multi,jia2020single}, learning for disentangled representations \cite{wang2020cross}, and learning to learn \cite{qin2020learning,shao2020regularized}. These methods almost implement DG on the complete representations from common modules ($i.e.$, CNN-BN-ReLU), but ignore fully taking advantage of subtle properties of global and local image statistics in FAS. Specifically, different normalization approaches lay stress on different statistics information in FAS. For example, Batch Normalization (BN) \cite{ioffe2015batch} based structures are usually used to summarize global image statistics, such as semantic features and physical attributes. Instance Normalization (IN) \cite{ulyanov2016instance} based structures focus on the specific sample for distinctive characteristics, such as liveness-related texture and domain-specific external factors. Thus, to mine different statistics in FAS, \cite{liu2021adaptive} adopts an adaptive approach to adjust the ratio of IN and BN in feature extraction. Differently, we adopt BN and IN based structures to separate the complete representation into global and local image statistics, denoted as content and style features respectively, then implement specific measures on them for generalizable FAS.

Besides, style transfer \cite{huang2017arbitrary} can be used to reassemble the pairs of content features as global statistics and style features as local statistics to form stylized features for specific supervision. As shown in Fig. \ref{fig:fig_head}, spoofing cues as style input can be applied to live faces to generate the corresponding spoof manipulations. Thus, \cite{laurensi2019style,yang2021few} directly utilize this approach for data augmentation before the training in FAS. However, these two-stage methods are inefficient in large-scale training. Thus, an end-to-end approach is adopted based on style transfer at the feature level in this work.

Combined with the abovementioned viewpoints, we propose a novel framework, called shuffled style assembly network (SSAN), based on style transfer at the feature level. Specifically, a two-stream structure is utilized to extract content and style features, respectively. For content information, they mainly record some global semantic features and physical attributes, thus a shared feature distribution is easily acquired by using adversarial learning. For style information, they preserve some discriminative information that is beneficial to enhance the distinction between living and spoofing. Different from the image-to-image style transfer proposed in \cite{huang2017arbitrary}, we stack up successive shuffled style assembly layers to reassemble various content and style features for a stylized feature space. Then, a contrastive learning strategy is adopted to enhance liveness-related style information and suppress domain-specific one. Lastly, our end-to-end architecture and training approach are more suitable for large-scale training in reality.


Due to the data distribution difference between academic and industrial scenarios, previous evaluation protocols are limited to reflect the genuine performance of algorithms in reality. Thus, to simulate the data quantity and distribution in reality, we combine twelve datasets to build a large-scale evaluation benchmark and further verify the effectiveness of algorithms. Specifically, the TPR@FPR at specific values as the metrics are utilized to evaluate the performance of different models on each dataset, where all live samples as negative cases and partial spoof samples as positive cases.


The main contributions of this work are four-fold:

$\bullet$ To utilize the global and local statistics separately for their unique properties, we propose a novel architecture called shuffled style assembly network (SSAN) for generalizable face anti-spoofing.

$\bullet$ To enhance liveness-relative style information and suppress domain-specific one, we adopt a contrastive learning approach to control the stylized features close or far from the anchor feature. The corresponding loss function is utilized to supervise our network.

$\bullet$ Based on the real-world data distribution, we combine twelve public datasets into a large-scale benchmark for face anti-spoofing in reality. The metric of single-side TPR@FPR is proposed for a comprehensive assessment.

$\bullet$ Our proposed methods achieve the state-of-the-art performance on existing and proposed benchmarks.

\section{Related Work}

\textbf{Face Anti-Spoofing.} Traditional methods usually extract hand-crafted features such as LBP \cite{de2012lbp} and SIFT \cite{patel2016secure} to split living and spoofing. In the era of deep learning, \cite{yang2014learn} trains CNNs to learn a binary classifier. Auxiliary information such as depth map \cite{atoum2017face}, reflection map \cite{yu2020face}, and rPPG \cite{lin2019face} is utilized to explore additional details for FAS.


To make the algorithm generalize well to unseen scenarios, domain adaptation (DA) and domain generation (DG) techniques are developed. \cite{li2018unsupervised} minimizes MMD \cite{gretton2012kernel} to pull close between different distributions. \cite{wang2019improving} leverages adversarial domain adaptation to learn a shared embedding space. \cite{shao2019multi} utilizes multiple domain discriminators to learn a generalized feature space. \cite{jia2020single} forms single-side adversarial learning to further improve the performance. \cite{wang2020cross,zhang2020face} utilize disentangled representation learning to isolate the liveness-related features for classification. To obtain general learning, meta-learning based methods \cite{qin2020learning,shao2020regularized,chen2021generalizable,wang2021self,qin2021meta} are introduced and developed for regular optimization.

Different from previous DG methods, we split the complete representation into content and style ones with various supervision. Then, a generalized feature space is obtained by resembling features under a contrastive learning strategy.

\begin{figure*}
	\begin{center}
		\includegraphics[width=15cm]{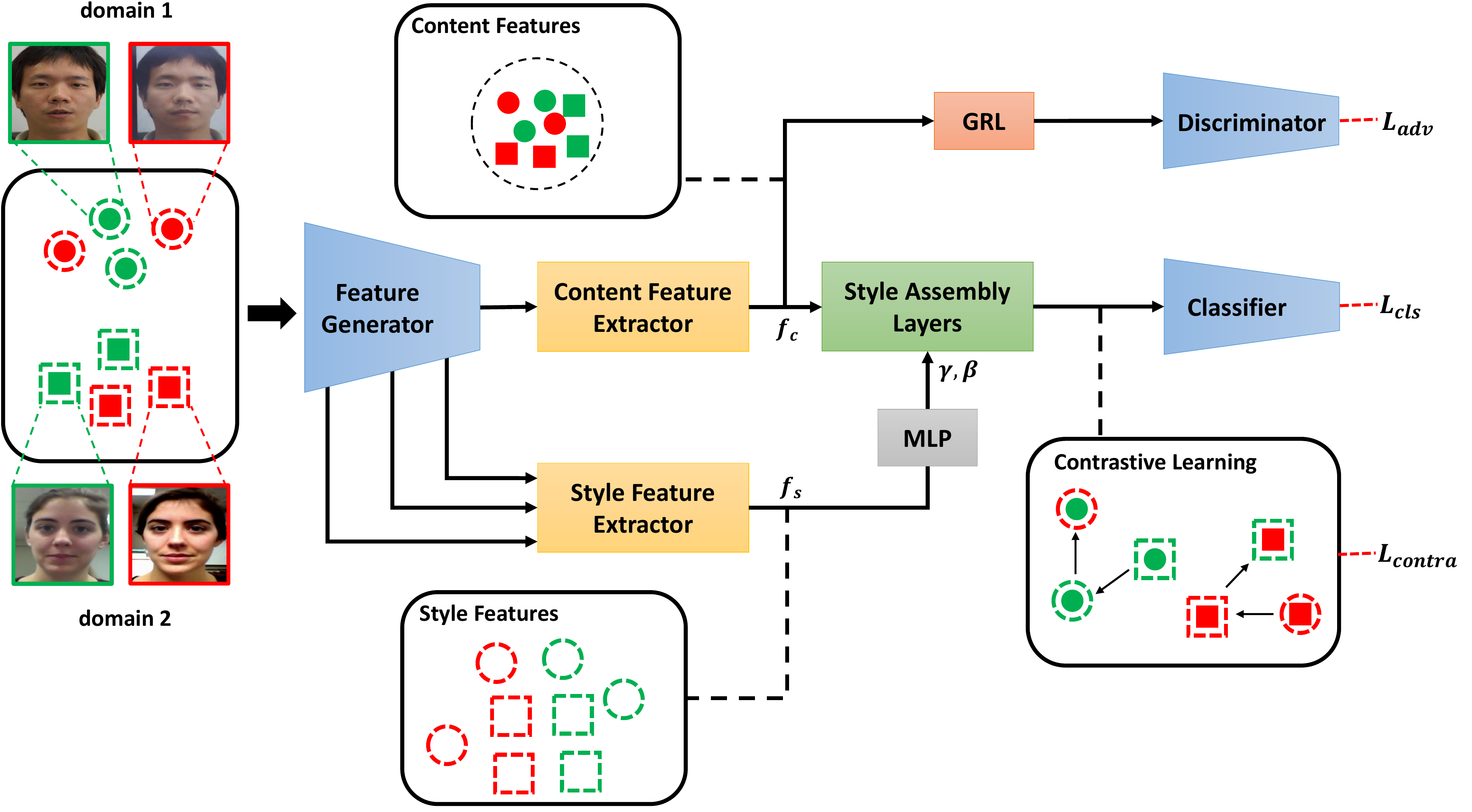}
	\end{center}
	\vspace{-16pt}
	\caption{The overall architecture of our shuffled style assembly network (SSAN). Firstly, RGB images from different domains are fed into the feature generator to obtain feature embeddings. Then, the feature extractor with GRL is trained to make the content feature indistinguishable for different domains by using adversarial learning. Meanwhile, another feature extractor collects multi-scale generated features to capture coarse-to-fine style information. Furthermore, to refine the style information related to FAS, a cascade of style assembly layers (SAL) are utilized to reassemble different content and style features when the corresponding contrastive learning strategy is designed.}
	\label{fig:img_main}
	\vspace{-8pt}
\end{figure*}

\textbf{Normalization and Style Transfer.} Normalization layers are essential in deep networks to eliminate covariate shifts and accelerate training. Batch Normalization (BN) \cite{ioffe2015batch} utilizes the statistics of the mini-batch to induce universal characteristics. Differently, Instance Normalization (IN) \cite{ulyanov2016instance} is proposed to exploit stylized characteristics for specific samples. Thus, the former lays stress on the global statistics and the latter emphasizes specific ones. \cite{huang2017arbitrary} proposes Adaptive Instance Normalization (AdaIN) for style transfer by utilizing target samples to control the scaling and shifting of source image normalized features. This style manipulation is widely used in generative tasks for texture synthesis \cite{nuriel2021permuted} and style transfer \cite{karras2019style}. Observing its effect on texture patterns, our method adopts this module to FAS. 

Different from the previous method \cite{liu2021adaptive,laurensi2019style,yang2021few} operating on normalization and image-level transformation, our method adopts AdaIN based layers to assemble different content and style features for a generalized feature space. 





\textbf{Protocols for Face Anti-Spoofing.} To evaluate the effectiveness of FAS methods, various protocols have been established, including intra-dataset intra-type protocol \cite{boulkenafet2017oulu,liu2018learning}, cross-dataset intra-type protocol \cite{li2018unsupervised}, intra-dataset cross-type protocol \cite{liu2019deep,george2019biometric}, and cross-dataset cross-type protocol \cite{arashloo2017anomaly,yu2020fas}. Especially, most protocols are merely constituted of single or double datasets, which may limit their evaluation capabilities for multiply data distributions. Thus, protocol OCIM \cite{shao2020regularized,shao2019multi} is used to evaluate their domain-generalization performance across multiple domains. 

Moreover, due to the limited amount of data, \cite{costa2019generalized} proposes an open-source framework to aggregate heterogeneous datasets for specific evaluation. Differently, we focus on the real-world data distribution, and more complex domain fields with different data distributions are obtained by fusing twelve different datasets including image and video formats. Thus, the merged dataset contains more sophisticated attack types, such as print, replay, mask, makeup, waxworks, $etc$. Besides, the evaluations under intra- and cross- domain scenarios among multiple datasets have been investigated by using the metric of single-side TPR@FPR, which is more suitable for realistic spectacles.


\section{Proposed Approach}


In this section, we introduce our shuffled style assembly network (SSAN) shown in Fig. \ref{fig:img_main}. Firstly, we present the two-stream part in our network for content and style information aggregation. Secondly, a shuffled style assembly approach is proposed to recombine various content and style features for a stylized feature space. Then, to suppress domain-specific style information and enhance liveness-related ones, contrastive learning is used in the stylized feature space. Lastly, the overall loss is integrated to optimize the network for stable and reliable training.

\subsection{Content and Style Information Aggregation}

Content information is usually represented by common factors in FAS, mainly including semantic features and physical attributes. Differently, style information describes some discriminative cues that can be divided into two parts in FAS tasks: domain-specific and liveness-related style information. Thus, content and style features are captured in the two-stream paths separately in our network. Specifically, the feature generator as a shallow embedding network captures multi-scale low-level information. Then, content and style feature extractors collect different image statistics by using specific normalization layers ($i.e.$, BN and IN).




For content information aggregation, we conjecture that small distribution discrepancies exist in different domains, based on the following facts: 1) Considering samples from various domains, they both contain facial areas, thus share a common semantic feature space; 2) Whether bona fide or attack presentation, their physical attributes such as shape and size are often similar. Therefore, we adopt adversarial learning to make generated content features indistinguishable for different domains. Specifically, the parameters of the content feature generator are optimized by maximizing the adversarial loss function while the parameters of the domain discriminator are optimized in the opposite direction. Thus, this process can be formulated as follows:
\begin{equation}
 \begin{split}
	&\underset{D}{min}\, \underset{G}{max}\ L_{adv}\left ( G,D \right )= \\
	& -\text{{\myfont E}}_{\left ( x,y \right )\sim\left ( X,Y_D \right )}\sum\nolimits_{i=1}^M\text{{\myfont 1}}\left [ i=y \right ]logD\left ( G\left ( x \right ) \right ),\\
\end{split}
\label{eqn:adv}
\end{equation}
where $Y_D$ is the set of domain labels and $M$ is the number of different data domains. $G$ and $D$ represent the content feature generator and domain discriminator, respectively. To optimize $G$ and the $D$ simultaneously, the gradient reversal layer (GRL) \cite{ganin2015unsupervised} is used to reverse the gradient by multiplying it by a negative scalar during the backward propagation.

For style information aggregation, we collect multi-layer features along with the hierarchical structure in a pyramid-like \cite{lin2017feature} approach, due to the different scales of style characteristics. For example, the brightness of scenes is mainly implicated in broad-scale features, while the texture of presentation materials usually focuses on local-scale regions.

\subsection{Shuffled Style Assembly}


Adaptive Instance Normalization (AdaIN) \cite{huang2017arbitrary} is an adaptive style transfer method, which can assemble a content input $x$ and a style input $y$, as follows:
\begin{equation}
\text{AdaIN}\left ( x,\gamma  ,\beta  \right )=\gamma  \left ( \frac{x-\mu (x)}{\sigma (x)} \right )+\beta ,
\end{equation}
where $\mu(\cdot)$ and $\sigma(\cdot)$ represent channel-wise mean and standard deviation respectively, $\gamma $ and $\beta$ are affine parameters generated from the style input $y$.

In this work, to combine content feature $f_c$ and style feature $f_s$, style assembly layers (SAL) are built up by using AdaIN layers and convolution operators with residual mapping, described as below:
\begin{equation}
 \begin{split}
  \gamma ,\beta &=\text{MLP}\left [ \text{GAP}\left ( f_s \right ) \right ], \\
  z&=\text{ReLU}\left [ \text{AdaIN}(K_1 \otimes f_c, \gamma ,\beta) \right ], \\
  \text{SAL}\left ( f_c,f_s \right )&=\text{AdaIN}(K_2\otimes z,\gamma ,\beta)+f_c, \\
\end{split}
\end{equation}
where $K_1$ and $K_2$ are $3\times3$ convolution kernels, $\otimes$ is the convolution operation, and $z$ is the intermediate variable.

However, $f_s$ contains not only liveness-related information, but also domain-specific one that may cause domain bias during network optimization. To alleviate this problem, the shuffled style assembly method is proposed to form auxiliary stylized features for domain generalization.




Given an input sequence of length $N$ in a mini-batch, $x_i$ represents the input sample, where $i\in \{1,2\dots N\}$. Its content feature can be expressed as $f_c(x_i)$ while the style feature as $f_s(x_i)$. Thus, the corresponding assembled feature space $S(x_i, x_i)$ can be formulated as follows:
\begin{equation}
S\left ( x_i,x_i \right )=\text{SAL}\left ( f_c\left ( x_i \right ),f_s\left ( x_i \right ) \right ),
\end{equation}
which represents the process of style assembly using paired content and style features of input sample $x_i$. Thus $S\left ( x_i,x_i \right )$ can be denoted as self-assembly features. 

Furthermore, to exploit liveness-related style features, we synthesize an auxiliary feature space by shuffling the original pairs of $f_c(x_i)$ and $f_s(x_i)$ randomly, as follows:
\begin{equation}
 \begin{split}
S\left ( x_i,x_{i^*} \right )&=\text{SAL}\left ( f_c\left ( x_i \right ),f_s\left ( x_{i^*} \right ) \right ), \\
i^*&\in random\left \{ 1,2,\dots,N \right \},
\end{split}
\end{equation}
where $random$ means a uniformly chosen permutation. $S\left ( x_i,x_{i^*} \right )$ can be denoted as shuffle-assembly features.


\subsection{Contrastive Learning for Stylized Features}

\begin{figure}
	\begin{center}
		\includegraphics[width=8cm]{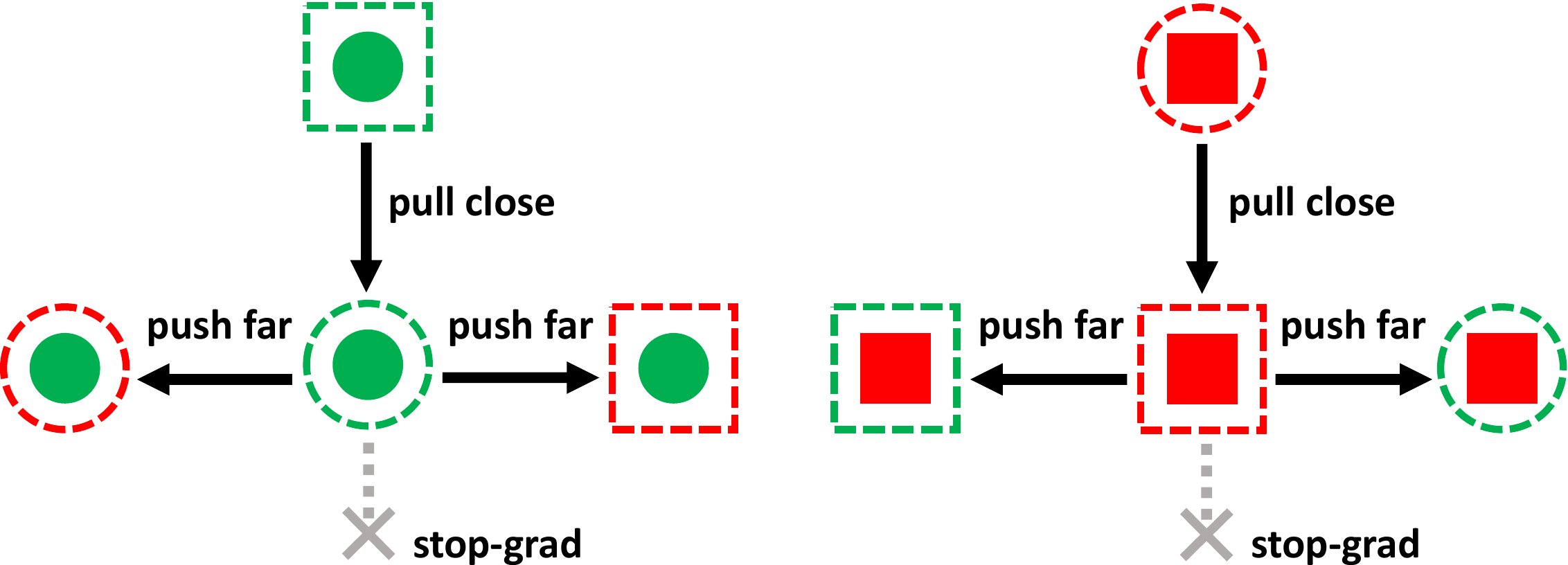}
	\end{center}
	\vspace{-15pt}
	\caption{The illustration of the contrastive learning between self-assembly and shuffle-assembly features. Different shapes represent data from different domains: $round$=domain 1, $square$=domain 2. Different colors represent different liveness information: $green$=living, $red$=spoofing. Lastly, the dotted line represents style information while the interior solid represents content information.}
	\label{fig:img_constra}
	\vspace{-8pt}
\end{figure}

From the view of style features, a major obstacle is that domain-specific style features may conceal liveness-related ones in cross-domain scenarios, which may cause mistakes in judgment. To solve this problem, we propose a contrastive learning approach to emphasize liveness-related style features as well as suppress domain-specific ones.


After combining content and style features, we obtain self-assembly features $S(x_i,x_i)$ and shuffle-assembly features $S(x_i,x_{i^*})$. For $S(x_i,x_i)$, we input them to the classifier and supervise them using our binary ground-turth signals with the loss function $L_{cls}$. For $S(x_i,x_{i^*})$, we measure their difference with $S(x_i,x_i)$ by using cosine similarity:
\begin{equation}
Sim\left ( a,b \right )=-\frac{a}{\left \| a \right \|_2}\cdot \frac{b}{\left \| b \right \|_2},
\end{equation}
where $\left \| \cdot  \right \|_2$ is $l_2$-norm, $a$ and $b$ represent two compared features. This is equivalent to the mean squared error of $l_2$-normalized vectors \cite{grill2020bootstrap}. 

As shown in Fig. \ref{fig:img_constra}, self-assembly features $S(x_i,x_i)$ are set as anchors in the stylized features space. Inspired by \cite{chen2021exploring}, a stop-gradient (stopgrad) operation is implemented on $S(x_i,x_i)$ to fix their position in the feature space. Then, the shuffle-assembly features $S(x_i,x_{i^*})$ are guided to go close or far toward their corresponding anchors $S(x_i,x_i)$ according to the liveness information. During the process, back-propagation is applied through the shuffle-assembly features but not through self-assembly ones, and the liveness-intensive style information is further aggregated. Thus, the contrastive loss $L_{contra}$ can be formulated as follows:
\begin{equation}
 L_{contra}=\sum\nolimits_{i=1}^NEq\left ( x_i,x_{i^*} \right )\cdot Sim\left ( \text{stopgrad}(a),b \right ),
 \label{eqn:contra}
 \end{equation}
 where $a=S(x_i,x_i)$ and $b=S(x_i,x_{i^*})$. $Eq(x_i,x_{i^*})$ measures the consistency of the liveness labels between $x_i$ and $x_{i^*}$, which can be formulated as follows:
 \begin{equation}
 Eq\left ( x_i,x_{i^*} \right )=\left\{
\begin{aligned}
&+1,\quad \text{label}(x_i)==\text{label}(x_{i^*}), \\
&-1,\quad \text{otherwise.}
\end{aligned}
\right.
 \end{equation}
Finally, The whole process of our framework can be described in Algorithm \ref{alg:1} in detail. 
\vspace{-3pt}
\begin{algorithm}
	\renewcommand{\algorithmicrequire}{\textbf{Input:}}
	\renewcommand{\algorithmicensure}{\textbf{Output:}}
	\caption{The optimization strategy of SSAN.}
	\label{alg:1}
	\begin{algorithmic}[1]
		\REQUIRE Mixture domain dataset $D_s$ = $\left \{ x_{i}^{s}, y_{i}^{s} \right \}_{i=1}^{n_s}$, initialized CNN model $\Phi_0(\cdot)$.
		\ENSURE Final CNN model parameter $\Phi_T(\cdot)$.
		\WHILE {not end of iteration}
		    \STATE Shuffle the input sequence for the permuted sequence $\left \{ x_{i^*}\mid i^*=random\left [ 1,2,\dots,N \right ] \right \}$.
		    \STATE Input $x_i$ for content feature $f_c(x_i)$ and style feature $f_s(x_i)$. Input $x_{i^*}$ for style feature $f_s(x_{i^*})$.
    		\STATE Input $f_c(x_i)$ to the discriminator and compute the adversarial loss $L_{adv}$ based on Eqn. (\ref{eqn:adv}).
    		\STATE Assemble $f_c(x_i)$ and $f_s(x_i)$ to get self-assembly features $S(x_i, x_i)$. Assemble $f_c(x_i)$ and $f_s(x_{i^*})$ to get shuffle-assembly features $S(x_i, x_{i^*})$.
    		\STATE Input $S(x_i, x_i)$ to the classifier and compute the classification loss $L_{cls}$.
    		\STATE Utilize $S(x_i, x_i)$ and $S(x_i, x_{i^*})$ to compute the contrastive loss $L_{contra}$ based on Eqn. (\ref{eqn:contra}).
    		\STATE Compute $L_{overall}=L_{cls}+\lambda_1\cdot L_{adv}+\lambda_2\cdot L_{contra}$. Make gradient back propagation and update the model parameters $\Phi(\cdot)$.
		\ENDWHILE
		\STATE Evaluate $\Phi_T(\cdot)$ on the testing data $D_t$.
	\end{algorithmic}  
\end{algorithm}
\vspace{-3pt}
 
\subsection{Loss Function}

After describing the operating of our network, we collect the overall loss function $L_{overall}$ for stable and reliable training, which can be formulated as follows:
\begin{equation}
L_{overall}=L_{cls}+\lambda_1\cdot L_{adv}+\lambda_2\cdot L_{contra},
\end{equation}
where $\lambda_1$ and $\lambda_2$ are two hyper-parameters to balance the proportion of different loss functions.





\begin{table}[htbp]
\footnotesize
  \centering
  \caption{The datasets and their corresponding numbers we use in the large-scale benchmark.}
  \vspace{-8pt}
    \begin{tabular}{|c|c|c|c|}
    \hline
    Dataset & Number & Dataset & Number \\
    \hline
    CASIA-MFSD \cite{zhang2012face} & D1    & Rose-Youtu \cite{li2018unsupervised} & D7 \\
    \hline
    REPLAY-ATTACK \cite{chingovska2012effectiveness} & D2    & WFFD \cite{jia20203d} & D8 \\
    \hline
    MSU-MFSD \cite{wen2015face} & D3    & CelebA-Spoof \cite{zhang2020celeba} & D9 \\
    \hline
    HKBU-MARs V2 \cite{yuen20193d} & D4    & CASIA-SURF \cite{zhang2019dataset} & D10 \\
    \hline
    OULU-NPU \cite{boulkenafet2017oulu} & D5    & WMCA \cite{george2019biometric} & D11 \\
    \hline
    SiW \cite{liu2018learning}  & D6    & CeFA \cite{liu2021casia} & D12 \\
    \hline
    \end{tabular}%
     \vspace{-12pt}
  \label{tab:data_num}%
\end{table}%

\section{Large-Scale FAS Benchmarks}
There exists a gap between academia and industry, which can be summarized as the following two aspects.

\textbf{Data Quantity.} Compared with the authentic scenarios, the amount of data in academia is still too small, which may cause overfitting of the model and limit the development of the algorithm. To overcome this problem, we merge twelve datasets then design corresponding intra- and inter- dataset testing protocols to further evaluate our method.

\begin{table*}[htbp]
\footnotesize
  \centering
  \vspace{-7pt}
  \caption{The results of cross-dataset testing on OULU-NPU, CASIA-MFSD, Replay-Attack, and MSU-MFSD.} 
  \vspace{-8pt}
    \begin{tabular}{|c|c|c|c|c|c|c|c|c|}
    \hline
    \multicolumn{1}{|c|}{\multirow{1}[4]{*}{Method}} & \multicolumn{2}{c|}{O\&C\&I to M} & \multicolumn{2}{c|}{O\&M\&I to C} & \multicolumn{2}{c|}{O\&C\&M to I} & \multicolumn{2}{c|}{I\&C\&M to O} \\
\cline{2-9}          & \multicolumn{1}{l|}{HTER(\%)} & \multicolumn{1}{c|}{AUC(\%)} & \multicolumn{1}{l|}{HTER(\%)} & \multicolumn{1}{c|}{AUC(\%)} & \multicolumn{1}{l|}{HTER(\%)} & \multicolumn{1}{c|}{AUC(\%)} & \multicolumn{1}{l|}{HTER(\%)} & \multicolumn{1}{c|}{AUC(\%)} \\
    \hline
    MMD-AAE \cite{li2018domain} & 27.08  & 83.19  & 44.59 & 58.29 & 31.58 & 75.18 & 40.98 & 63.08 \\
    MADDG \cite{shao2019multi} & 17.69 & 88.06 & 24.50  & 84.51 & 22.19 & 84.99 & 27.98 & 80.02 \\
    SSDG-M \cite{jia2020single} & 16.67 & 90.47 & 23.11  & 85.45 & 18.21 & 94.61 & 25.17 & 81.83 \\
    DR-MD-Net \cite{wang2020cross} & 17.02 & 90.10 & 19.68  & 87.43 & 20.87 & 86.72 & 25.02 & 81.47 \\
    RFMeta \cite{shao2020regularized} & 13.89  & 93.98  & 20.27 & 88.16 & 17.30  & 90.48 & 16.45 & 91.16 \\
    NAS-FAS \cite{yu2020fas} & 19.53 & 88.63 & 16.54 & 90.18 & 14.51 & 93.84 & \textbf{13.80}  & \textbf{93.43} \\
    D$^2$AM \cite{chen2021generalizable} & 12.70 & 95.66 & 20.98 & 85.58 & 15.43 & 91.22 & 15.27  & 90.87 \\
    SDA \cite{wang2021self} & 15.40 & 91.80 & 24.50 & 84.40 & 15.60 & 90.10 & 23.10  & 84.30 \\
    DRDG \cite{liudual} & 12.43 & 95.81 & 19.05 & 88.79 & 15.56 & 91.79 & 15.63  & 91.75 \\
    ANRL \cite{liu2021adaptive} & 10.83  & \textbf{96.75}  & 17.83  & 89.26 & 16.03 & 91.04 & 15.67 & 91.90 \\
    \textbf{SSAN-M (Ours)}   & \textbf{10.42} & 94.76 & \textbf{16.47} & \textbf{90.81} & \textbf{14.00} & \textbf{94.58} & 19.51 & 88.17 \\
    \hline
    SSDG-R \cite{jia2020single} & 7.38 & 97.17 & 10.44  & 95.94 & 11.71 & 96.59 & 15.61 & 91.54 \\
    \textbf{SSAN-R (Ours)}   & \textbf{6.67} & \textbf{98.75} & \textbf{10.00} & \textbf{96.67} & \textbf{8.88} & \textbf{96.79} & \textbf{13.72} & \textbf{93.63} \\
    \hline
    \end{tabular}%
    \vspace{-8pt}
  \label{tab:big_cross}%
\end{table*}%

\textbf{Data Distribution and Evaluation Metrics.} In real-world data distribution, live faces usually account for the majority. However, most existing evaluation protocols collect almost equivalent live and spoof faces as testing set to calculate their average error rate for evaluation, which may disagree with the reality. Besides, data in reality usually consists of multiple fields with different distributions. Nevertheless, academic datasets usually contain fewer data domains. To reduce the above inconsistencies, multiple datasets are used as training and testing sets simultaneously in our protocols.  Specifically, in the training stages, all of the training data are used to optimize our models. In the inferring stages, due to the similar distribution of live faces\cite{jia2020single}, we gather all live data from each testing dataset as the negative cases, then partial spoof data in the current testing dataset is arranged as positive cases. Lastly, the mean and variance of true positive rate (TPR) of false-positive rate (FPR) are computed along with each testing dataset for an overall evaluation.

Twelve datasets are used in the large-scale FAS Benchmarks, which are numbered as shown in Table \ref{tab:data_num}. The evaluation protocols are designed as follows:

$\bullet$ \textbf{Protocol 1.} This protocol is implemented in an intra-dataset evaluation scenario. Specifically, all datasets are used as training and testing sets, simultaneously.

$\bullet$ \textbf{Protocol 2.} This protocol is implemented in a cross-domain evaluation scenario by dividing these datasets into two piles: $P1$: \{D3, D4, D5, D10, D11, D12\}, $P2$: \{D1, D2, D6, D7, D8, D9\}. Thus, there contain two sub-protocols: \textbf{protocol 2\_1}: training on $P1$ and testing on $P2$; \textbf{Protocol 2\_2}: training on $P2$ and testing on $P1$. Note that the cross-domain protocols are more challenging as the testing set covers more unseen datasets and more complex unknown attacks, which are correlated to real-world scenarios.



More details are provided in supplementary materials.


\section{Experiments}



\subsection{Implementation Details}

\textbf{Data Preparation.}
The datasets shown in Table \ref{tab:data_num} contain image and video data. For image data, we utilize all images of them. For video data, we extract frames of them at specific intervals. After obtaining data in image format, we adopt MTCNN \cite{zhang2016joint} for face detection, then crop and resize faces to $256\times 256$ as RGB input. Moreover, a dense face alignment approach ($i.e.$, PRNet \cite{feng2018joint}) is used to generate the ground-truth depth maps with size $32 \times 32$ for genuine faces, while spoof depth maps are set to zeros.


\textbf{Networks Setting.} Similar to \cite{jia2020single}, two structures are established, denoted as SSAN-M and SSAN-R. Specifically, SSAN-M adopts the embedding part of DepthNet \cite{liu2018learning} while SSAN-R adopts that of ResNet-18 \cite{he2016deep} for feature generation. More details are in supplementary materials.



\textbf{Training Setting.} Due to the limit of the GPU memory size, the batch size is set to 16 for SSAN-M and set to 256 for SSAN-R. Different ground-turth are used as supervision signals: depth maps for SSAN-M and binary labels for SSAN-R. Therefore, their corresponding $L_{cls}$ are mean-squared and cross-entropy loss, respectively. $\lambda_1$ and $\lambda_2$ are set to 1 in training. The Adam optimizer with the learning rate (lr) of 1e-4 and weight decay of 5e-5 is used in the experiments on OCIM. The SGD optimizer with the momentum of 0.9 and weight decay of 5e-4 is used in the experiments on proposed protocols. Its initial lr is 0.01 and decreases by 0.2 times every two epochs until the $30^{th}$ epoch.


\textbf{Testing Setting.} In testing, we calculate the final classification score to separate bona fide and attack presentations. Specifically, the mean value of the predicted depth map is the final score for SSAN-M, while the value of the sigmoid function on living is the final score for SSAN-R.

\subsection{Experiment on OCIM.}

Four datasets are used to evaluate the performance of SSAN in different cross-domain scenarios following the implementation of \cite{shao2019multi}: OULU-NPU \cite{boulkenafet2017oulu} (O), CASIA-MFSD \cite{zhang2012face} (C), Replay-Attack \cite{chingovska2012effectiveness} (I), and MSU-MFSD \cite{wen2015face} (M).






\textbf{Experiment in Leave-One-Out (LOO) Setting.} For an overall evaluation, we conduct cross-dataset testing by using the LOO strategy: three datasets are selected for training, and the rest one for testing. We compare our models with the recent SOTA methods, as shown in Table \ref{tab:big_cross}. It can be observed that our SSAN-M shows the best performance on protocols of O\&C\&I to M, O\&M\&I to C, O\&C\&M to I, and the competitive performance on the protocol of I\&C\&M to O. These results demonstrate the domain generalization capacity of our method. Moreover, when we adopt the ResNet18-based network denoted as SSAN-R, its performance obtains an excellent improvement and exceeds the model SSDG-R proposed in \cite{jia2020single} with similar settings. The above phenomenon indicates our network SSAN-R is more effective in the cross-dataset scenario, thus will be further measured in the large-scale protocols we propose.



\begin{table}[htbp]
\footnotesize
  \centering
  \vspace{-8pt}
  \caption{Comparison results on limited source domains.}
  \vspace{-10pt}
  \setlength{\tabcolsep}{1.8mm}{
  \begin{tabular}{|c|c|c|c|c|}
    \hline
    \multirow{1}[4]{*}{Method} & \multicolumn{2}{c|}{M\&I to C} & \multicolumn{2}{c|}{M\&I to O} \\
\cline{2-5}          & HTER(\%)  & AUC(\%)   & HTER(\%)  & AUC(\%) \\
    \hline
    MS-LBP \cite{maatta2011face} & 51.16 & 52.09 & 43.63 & 58.07 \\
    IDA \cite{wen2015face}   & 45.16 & 58.80  & 54.52 & 42.17 \\
    LBP-TOP \cite{de2014face} & 45.27 & 54.88 & 47.26 & 50.21 \\
    MADDG \cite{shao2019multi} & 41.02 & 64.33 & 39.35 & 65.10 \\
    SSDG-M \cite{jia2020single} & 31.89 & 71.29 & 36.01 & 66.88 \\
    DR-MD-Net \cite{wang2020cross} & 31.67 & 75.23 & 34.02 & 72.65 \\
    ANRL \cite{liu2021adaptive}  & 31.06 & 72.12 & 30.73 & 74.10 \\
    \textbf{SSAN-M (Ours)} &  \textbf{30.00}     &  \textbf{76.20}     &   \textbf{29.44}    & \textbf{76.62} \\
    \hline
    \end{tabular}%
    \vspace{-5pt}
  }
  \label{tab:limit}%
\end{table}%

\textbf{Experiment on Limited Source Domains.} We also evaluate our method when extremely limited source domains are available. Specifically, MSU-MFSD and Replay-Attack are selected as the source domains for training and the remaining two ($i.e.$, CASIA-MFSD and OULU-NPU) will be used as the target domains for testing respectively. As shown in Table \ref{tab:limit}, our method achieves the lowest HTER and the highest AUC despite limited source data, which proves the modeling efficiency and generalization capability of our network in a challenging task.

\subsection{Experiment on Proposed Benchmarks} To further evaluate the performance of our method in reality, we conduct the experiments on the large-scale FAS benchmark we proposed, as shown in Table \ref{tab:large_sacle}. Different network structures ($i.e.$, CNN \cite{he2016deep} and Transformer \cite{touvron2021training}) and some recent SOTA methods ($i.e.$, CDCN \cite{yu2020searching} and SSDG \cite{jia2020single}) are also conducted in their default settings for comparison. From the evaluation results, we can observe that our method achieves the best performance, exceeding that of other compared methods, which proves the effectiveness of our SSAN in real-world data distribution. It is worth noting that some methods have achieved excellent performance on existing protocols, but may suffer an acute degeneration in the large-scale benchmarks. This phenomenon further reveals the mismatch between academia and industry in FAS. More detailed analyses are in supplementary materials.

\begin{table}[htbp]
\footnotesize
  \centering
  \vspace{-6pt}
  \caption{The results on the large-scale FAS benchmarks.}
  \vspace{-8pt}
  \setlength{\tabcolsep}{1.2mm}{
    \begin{tabular}{|c|c|c|c|c|}
    \hline
    \multirow{1}[4]{*}{Prot.} & \multicolumn{1}{c|}{\multirow{1}[4]{*}{Method}} & \multicolumn{3}{c|}{TPR@FPR(\%)} \\
\cline{3-5}          &       & 10\%    & 1\%    & 0.1\% \\
    \hline
    \multirow{5}[1]{*}{1} & ResNet18 \cite{he2016deep} & 96.04$\pm$11.96      & 89.32$\pm$26.08      & 69.10$\pm$34.34 \\
          & Deit-T \cite{touvron2021training} &  97.75$\pm$5.70     & 90.38$\pm$16.08      & 73.42$\pm$30.00 \\
          & CDCN \cite{yu2020searching}  &   92.59$\pm$15.99    &  84.40$\pm$31.93     & 71.54$\pm$32.05 \\
          & SSDG-R \cite{jia2020single}  &   96.48$\pm$10.37    & 89.13$\pm$25.59      & 68.12$\pm$39.12 \\
          & \textbf{SSAN-R (Ours)} & \textbf{98.31$\pm$4.19}      & \textbf{90.51$\pm$22.31}      & \textbf{78.45$\pm$31.98} \\
    \hline
    \multirow{5}[1]{*}{2\_1} & ResNet18 \cite{he2016deep} &  55.64$\pm$22.05     &  17.53$\pm$13.44     & 3.64$\pm$3.93 \\
          & Deit-T \cite{touvron2021training} & 44.03$\pm$17.77      & 10.15$\pm$6.08      & 1.25$\pm$1.04 \\
          & CDCN \cite{yu2020searching}  &  55.92$\pm$21.45     & 11.07$\pm$8.21      & 0.69$\pm$0.74 \\
          & SSDG-R \cite{jia2020single}  &  53.44$\pm$19.23     &   3.27$\pm$3.09    & 0.06$\pm$0.06 \\
          & \textbf{SSAN-R (Ours)} &   \textbf{63.61$\pm$21.69}    & \textbf{25.56$\pm$18.07}       & \textbf{6.58$\pm$5.56} \\
    \hline
    \multirow{5}[1]{*}{2\_2} & ResNet18 \cite{he2016deep} &    63.38$\pm$27.54   & 41.53$\pm$30.41      & 19.00$\pm$14.79 \\
          & Deit-T \cite{touvron2021training} &     63.29$\pm$13.39  & 30.46$\pm$19.15      & 11.30$\pm$9.45 \\
          & CDCN \cite{yu2020searching}  &  20.97$\pm$25.23     &  3.58$\pm$4.83     & 0.58$\pm$0.88 \\
          & SSDG-R \cite{jia2020single}  & 41.13$\pm$28.45      & 7.19$\pm$8.73      & 1.94$\pm$2.35 \\
          & \textbf{SSAN-R (Ours)} &   \textbf{64.54$\pm$28.36}    & \textbf{47.07$\pm$33.71}      & \textbf{31.61$\pm$23.33} \\
    \hline
    \end{tabular}%
    \vspace{-8pt}
    }
  \label{tab:large_sacle}%
\end{table}%


\subsection{Ablation Study}
To verify the superiority of our SSAN as well as the contributions of each component, multiple incomplete models are built up by controlling different variables. All results are measured in the same manner, as shown in Table \ref{tab:ablation}.




\textbf{Effectiveness of Different Components.} To verify the effectiveness of generalized content feature space, we conduct the experiments of SSAN w/o $L_{adv}$. Specifically, content features usually record some common patterns in FAS, which is easier to reduce their domain difference, compared to directly operating on the complete features. Besides, to make assembly between arbitrary combinations of content and style features for domain generalization, stripping domain distinction from content information is indispensable.

On the other hand, to prove the importance of contrastive learning for shuffled stylized features, the experiments of SSAN w/o $L_{contra}$ are implemented for comparison. The quantitative results indicate that the style assembly guided by liveness-intensive cues is beneficial to improve the performance for cross-domain FAS tasks.

\begin{figure}
	\begin{center}
		\includegraphics[width=8.0cm]{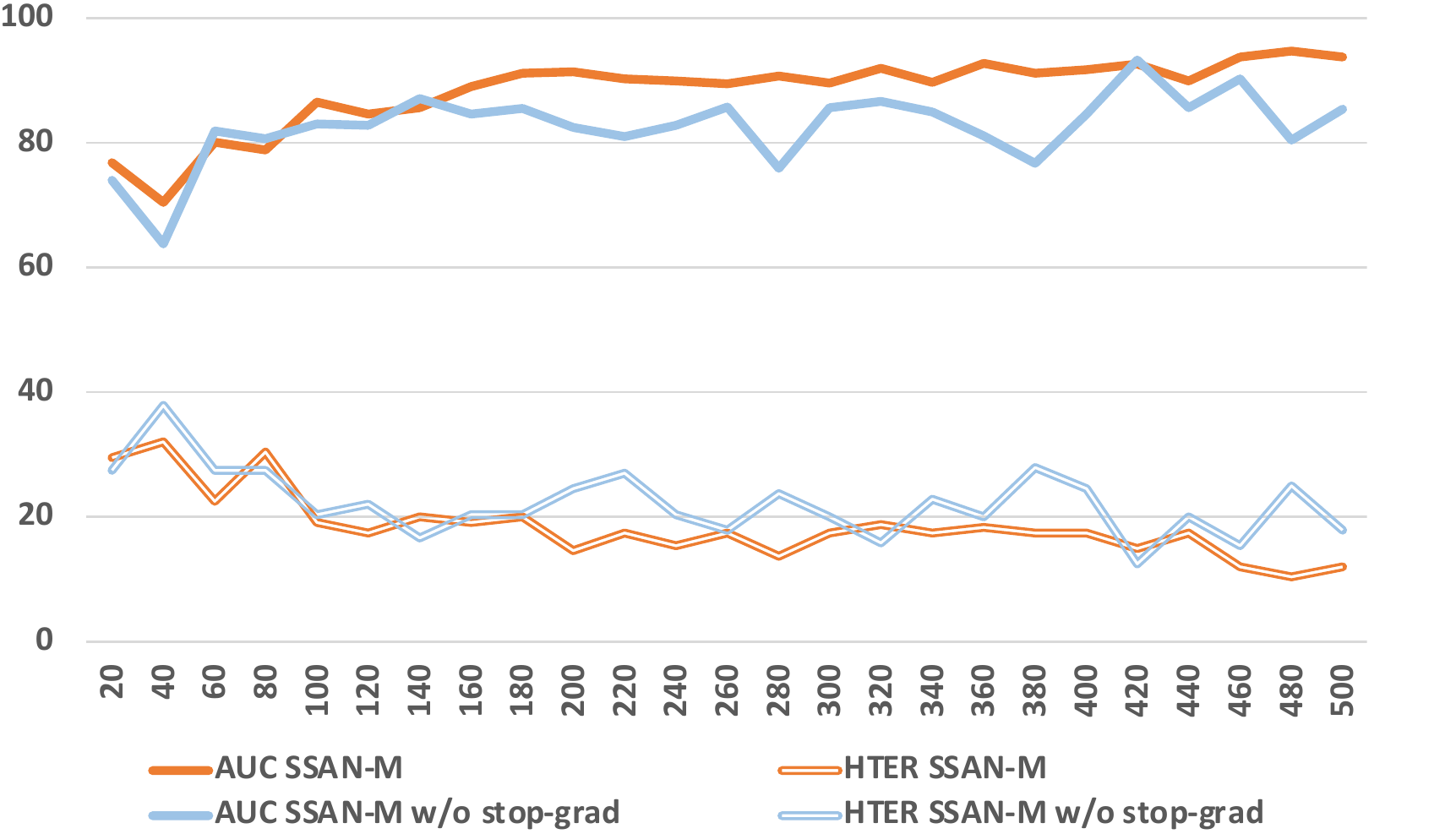}
	\end{center}
	\vspace{-15pt}
	\caption{The comparison curves between SSAN-M and SSAN-M w/o stop-grad under protocol O\&C\&I to M. The $x$-axis represents the number of epochs while the $y$-axis records the value of AUC(\%) the HTER(\%), as shown in the legend.}
	\label{fig:curves}
	\vspace{-12pt}
\end{figure}

\textbf{Impact of the Stop-Gradient Operation.} In contrastive learning for stylized features, the self-assembly features adopt the approach of stop-gradient to fix their position in the feature space as an anchor. Then, their corresponding shuffle-assembly features obeying on the liveness information to go close or far toward them. The ablation experiment of SSAN w/o stop-grad shows its effectiveness of feature aggregation in contrastive learning for emphasizing liveness-related style information and suppressing domain-specific ones. Besides, from the continuous evaluation curves shown in Fig. \ref{fig:curves}, it can be summarized that the stop-gradient operation will contribute to stable training.


\textbf{Comparison Between the Hard and Soft Supervision.} The relative movement approach in contrastive learning we adopt can be regarded as soft supervision in stylized feature space, compared to the direct supervision using the ground-turth. To investigate their different efficiency, we conduct the experiment of w/ hard-sup for an ablation study between them, as shown in Table \ref{tab:ablation}. The declining performance shows the soft supervision method is more suitable for our networks under the cross-domain testing scenarios.

\textbf{Analysis of Contrastive Learning.} Existing works \cite{liu2021contrastive,mishra2021improved} implement classical supervised contrastive learning (SCL) on the complete representation in FAS. Differently, our method conducts contrastive learning between self-assembly and shuffle-assembly features. To make a comparison between them, the experiment of w/ SCL is conducted by implementing contrastive learning on self-assembly directly. The final results demonstrate the efficiency of the auxiliary features in contrastive learning, which are built in a shuffle-then-assembly approach.
 
\begin{table*}[htbp]
\footnotesize
  \centering
  \vspace{-4pt}
  \caption{Evaluations of different components of the proposed method with different architectures.} 
   \vspace{-8pt}
    \begin{tabular}{|c|c|c|c|c|c|c|c|c|}
    \hline
    \multicolumn{1}{|c|}{\multirow{1}[4]{*}{Method}} & \multicolumn{2}{c|}{O\&C\&I to M} & \multicolumn{2}{c|}{O\&M\&I to C} & \multicolumn{2}{c|}{O\&C\&M to I} & \multicolumn{2}{c|}{I\&C\&M to O} \\
\cline{2-9}          & \multicolumn{1}{l|}{HTER(\%)} & \multicolumn{1}{c|}{AUC(\%)} & \multicolumn{1}{l|}{HTER(\%)} & \multicolumn{1}{c|}{AUC(\%)} & \multicolumn{1}{l|}{HTER(\%)} & \multicolumn{1}{c|}{AUC(\%)} & \multicolumn{1}{l|}{HTER(\%)} & \multicolumn{1}{c|}{AUC(\%)} \\
    \hline
    SSAN-M w/o $L_{adv}$  & 10.42 & \textbf{94.83} & 24.44 & 81.60 & 24.75  & 83.01 & 27.11 & 80.41 \\
    SSAN-M w/o $L_{contra}$  & 12.50 & 93.59 & 17.59 & 89.33 & 14.75  & 92.67 & 22.47 & 85.79 \\
    SSAN-M w/o stop-grad  & 12.50 & 93.33 & 20.93 & 85.02 & 16.38  & 89.78 & 23.65 & 83.14 \\
    SSAN-M w/ hard-sup  & 12.08 & 93.42 & 28.89 & 77.70 & 20.61  & 86.46 & 24.83 & 82.39 \\
    SSAN-M w/ SCL  & 12.92 & 92.50 & 23.70 & 84.67 & 18.75  & 87.28 & 25.45 & 82.03 \\
    \textbf{SSAN-M (Ours)}   & \textbf{10.42} & 94.76 & \textbf{16.67} & \textbf{90.81} & \textbf{14.00} & \textbf{94.58} & \textbf{19.51} & \textbf{88.17} \\
    \hline
    SSAN-R w/o $L_{adv}$  & 10.83 & 94.08 & 14.26 & 94.48 & 12.25  & 94.93 & 14.27 & 92.83 \\
    SSAN-R w/o $L_{contra}$  & 12.08 & 95.62 & 12.59 & 94.97 & 10.75  & 95.01 & 15.31 & 92.31 \\
    SSAN-R w/o stop-grad  & 11.25 & 93.46 & 11.30 & 95.11 & 9.00  & 96.03 & 14.06 & 93.14 \\
    SSAN-R w/ hard-sup  & 11.67 & 96.04 & 14.63 & 94.65 & 11.38  & 94.61 & 15.21 & 92.97 \\
    SSAN-R w/ SCL  & 11.25 & 94.00 & 12.04 & 94.91 & 12.50  & 95.34 & 15.80 & 92.95 \\
    \textbf{SSAN-R (Ours)}   & \textbf{6.67} & \textbf{98.75} & \textbf{10.00} & \textbf{96.67} & \textbf{8.88} & \textbf{96.79} & \textbf{13.72} & \textbf{93.63} \\
    \hline
    \end{tabular}%
  \label{tab:ablation}%
\end{table*}%

\subsection{Visualization and Analysis}

\begin{figure*}
    \vspace{-5pt}
	\begin{center}
		\includegraphics[width=15.8cm]{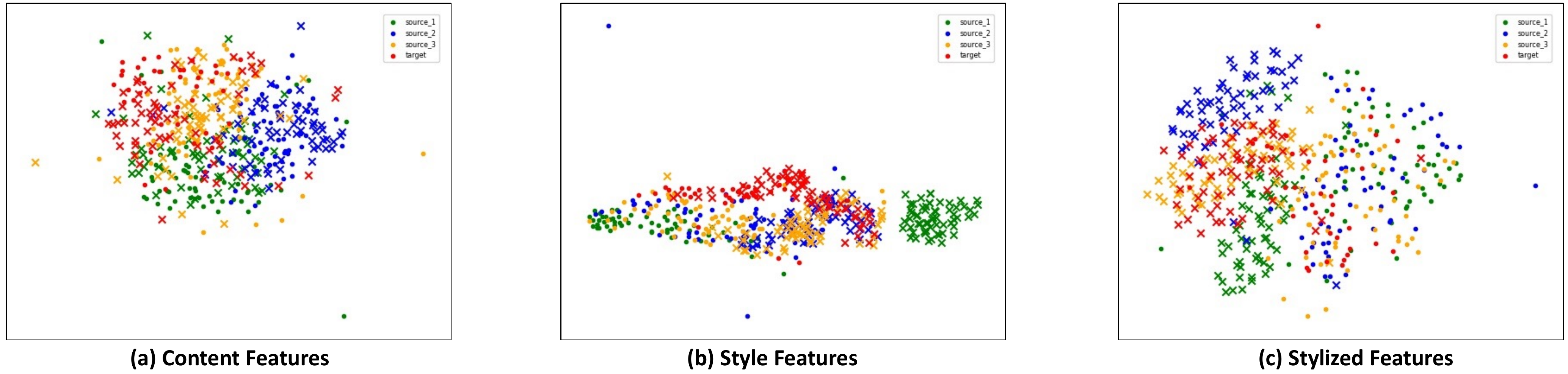}
	\end{center}
	\vspace{-15pt}
	\caption{The t-SNE \cite{van2008visualizing} visualization of different features under protocol O\&C\&I to M. The graphs of (a), (b), and (c) describe the feature distribution of content features, style features, and stylized features, respectively. Different colors indicate features from different domains: $green$=O, $blue$=C, $yellow$=I, $red$=M. Different shapes represent different liveness information: $point$=living, $cross$=spoofing.}
	\label{fig:tsne}
	\vspace{-15pt}
\end{figure*}

\textbf{Features Visualization.} To analyze the feature space learned by our SSAN method, we visualize the distribution of different features using t-SNE \cite{van2008visualizing}, as shown in Fig. \ref{fig:tsne}. For content features, it can be observed that their distribution is more compact and mixed, though they may belong to multiple databases or various liveness attributions. For style features, there exists a coarse boundary between living and spoofing along with a narrow distribution, despite no direct supervision on them. This phenomenon indicates that our contrastive learning for stylized features is effective to emphasize liveness-related style features as well as suppress other irrelevant ones, such as domain-specific information. For stylized features, we combine the content and style information for the classification between living and spoofing. The visualization results show that even though encountering an unknown distribution, our method still can generalize well to the target domain.

\begin{figure}
	\begin{center}
		\includegraphics[width=8.0cm]{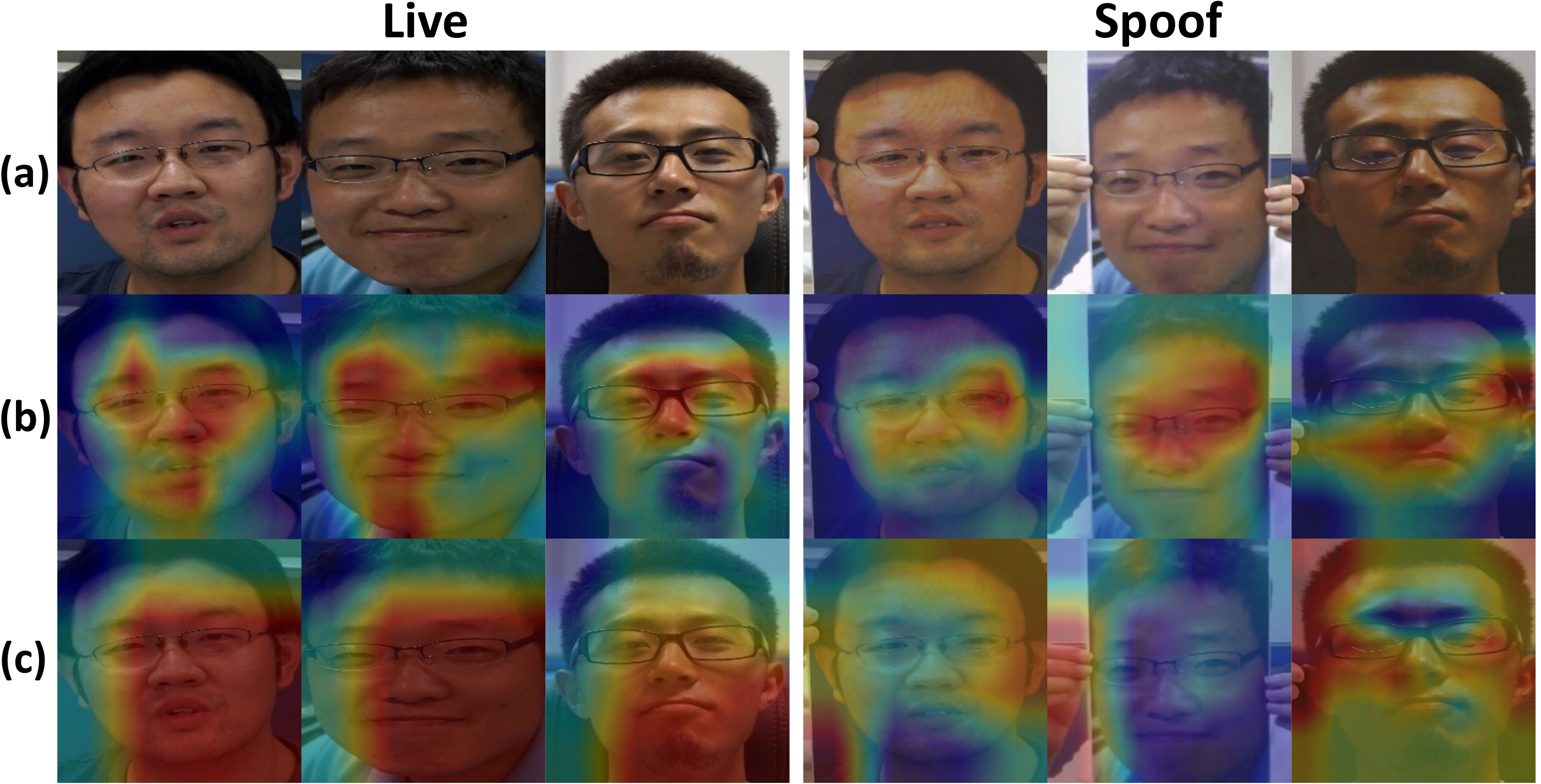}
	\end{center}
	\vspace{-16pt}
	\caption{Grad-CAM \cite{zhou2016cvpr} visualizations of activation areas under protocol O\&M\&I to C. (a): Original images. (b): Visualizations for content features generation. (c): Visualizations for assembled features (content + style) generation.}
	\label{fig:cam}
	\vspace{-15pt}
\end{figure}

\textbf{Attention Visualization.} To find the regions that led to content feature extraction and liveness detection, we adopt the Grad-CAM \cite{zhou2016cvpr} to describe their activation maps upon the original images, as shown in Fig. \ref{fig:cam}. It can be observed that despite living and spoofing, their content features both mainly focus on the landmark areas in faces that contain abundant semantic features and physical attributes. Then, after combined with the style information, the stylized features for classification show different activation properties: (1) For the live faces, our model lays the stress on the face regions to seek cues for judgment; (2) For the spoofing faces, some spoofing cues will be concentrated by our method, such as the moire phenomenon in replay attacks and the photo cut position in print attacks.

\section{Conclusion}
In this paper, we have proposed a novel shuffled style assembly network (SSAN) for generalizable face anti-spoofing (FAS). Different from the previous methods implemented on the complete features, we operate on content and style features separately due to their various properties. For content features, adversarial learning is adopted to make them domain-indistinguishable. For style features, a contrastive learning strategy is used to emphasize liveness-related style information while suppress domain-specific one. Then, the correct pairs of content and style features are reassembled for classification. Moreover, to bridge the gap between academia and industry, a large-scale benchmark for FAS is built up by aggregating existing datasets. Experimental results on existing and proposed benchmarks have demonstrated the superiority of our methods.

\textbf{Acknowledgements} This work was partially supported by the National Natural Science Foundation of China under Grants No. 61871052 and 62192784.



\clearpage
{\small
\bibliographystyle{ieee_fullname}
\bibliography{egbib}
}

\end{document}